\documentclass[sigconf]{acmart}
\renewcommand\footnotetextcopyrightpermission[1]{}
\usepackage{multirow}
\usepackage{colortbl}
\usepackage{makecell}
\AtBeginDocument{%
  }

\begin{document}
\title{Locate and Verify: A Two-Stream Network for Improved Deepfake Detection}

\author{Chao Shuai}
\orcid{0000-0002-4858-4602}
\affiliation{%
  \institution{Zhejiang University\&ZJU-Hangzhou Global Scientific and Technological Innovation Center}
  \city{Hangzhou}
  \state{Zhejiang}
  \country{China}
  \postcode{310058}
}
\email{chashuai@zju.edu.cn}

\author{Jieming Zhong}
\orcid{0009-0007-1933-9775}
\affiliation{%
  \institution{Zhejiang University}
  \city{Hangzhou}
  \state{Zhejiang}
  \country{China}
  \postcode{310058}
}
\email{jiemingzhong@zju.edu.cn}

\author{Shuang Wu}
\orcid{0000-0002-7551-7712}
\affiliation{%
  \institution{Black Sesame Technologies}
  \city{}
  \state{}
  \country{Singapore}
  \postcode{}
}
\email{wushuang@outlook.sg}

\author{Feng Lin}
\orcid{0000-0001-5240-5200}
\affiliation{%
  \institution{Zhejiang University}
  \city{Hangzhou}
  \state{Zhejiang}
  \country{China}
  \postcode{310058}
}
\email{flin@zju.edu.cn}

\author{Zhibo Wang}
\orcid{0000-0002-5804-3279}
\affiliation{%
  \institution{Zhejiang University}
  \city{Hangzhou}
  \state{Zhejiang}
  \country{China}
  \postcode{310058}
}
\email{zhibowang@zju.edu.cn}

\author{Zhongjie Ba}
\authornotemark[1]
\orcid{0000-0003-0921-8869}
\affiliation{%
  \institution{Zhejiang University}
  \city{Hangzhou}
  \state{Zhejiang}
  \country{China}
  \postcode{310058}
}
\email{zhongjieba@zju.edu.cn}

\author{Zhenguang Liu}
\authornote{Corresponding Authors: Zhenguang Liu, Zhongjie Ba.}
\orcid{0000-0002-7981-9873}
\affiliation{%
  \institution{Zhejiang University}
  \city{Hangzhou}
  \state{Zhejiang}
  \country{China}
  \postcode{310058}
}
\email{liuzhenguang2008@gmail.com}

\author{Lorenzo Cavallaro}
\orcid{0000-0002-3878-2680}
\affiliation{%
  \institution{University College London \& Zhejiang University}
  \city{London}
  \country{United Kingdom}
  \postcode{WC1E6BT}
}
\email{l.cavallaro@ucl.ac.uk}

\author{Kui Ren}
\orcid{0000-0003-3441-6277}
\affiliation{%
  \institution{Zhejiang University}
  \city{Hangzhou}
  \state{Zhejiang}
  \country{China}
  \postcode{310058}
}
\email{kuiren@zju.edu.cn}

\renewcommand{\shortauthors}{Chao Shuai et al.}




\begin{abstract}
  \par Deepfake has taken the world by storm, triggering a trust crisis. Current deepfake detection methods are typically inadequate in generalizability, with a tendency to overfit to image contents such as the background, which are frequently occurring but relatively unimportant in the training dataset. Furthermore, current methods heavily rely on a few dominant forgery regions and may ignore other equally important regions, leading to inadequate uncovering of forgery cues.
  \par In this paper, we strive to address these shortcomings from three aspects: (1) We propose an innovative two-stream network that effectively enlarges the potential regions from which the model extracts forgery evidence. (2) We devise three functional modules to handle the multi-stream and multi-scale features in a collaborative learning scheme. (3) Confronted with the challenge of obtaining forgery annotations, we propose a Semi-supervised Patch Similarity Learning strategy to estimate patch-level forged location annotations. Empirically, our method demonstrates significantly improved robustness and generalizability, outperforming previous methods on six benchmarks, and improving the frame-level AUC on Deepfake Detection Challenge preview dataset from 0.797 to 0.835 and video-level AUC on CelebDF$\_$v1 dataset from 0.811 to 0.847. Our implementation is available at \href{https://github.com/sccsok/Locate-and-Verify}{https://github.com/sccsok/Locate-and-Verify}.
\end{abstract}



\maketitle

\section{Introduction}
Over the past decade, we have witnessed the success of deep learning in various fields~\cite{CVPR2021DualConsecutiveNetwork, wang2023bi, yang2021deconfounded, tolosana2020deepfakes, yin2022mix}, especially deepfake technology standing out as a prominent catalyst for stimulating creative expression. However, this technology's accessibility, facilitated by numerous off-the-shelf tools like Face2Face, FSGAN, and SimSwap~\cite{face2face, fsgan, simswap}, has also given rise to concerns about the misuse of creating fake videos fabricating people's words and actions~\cite{synthesizing, liu2022copy, prajwal2020lip}. For example, in March 2022, hackers created a fake video of Ukrainian President Zelenskyy delivering a speech urging soldiers to surrender, and during the U.S. presidential election, a deepfake video of former President Obama provoking presidential candidate Trump was circulated. These incidents are far from mere curiosities, and their potential sociopolitical and security implications are too significant to overlook. Undoubtedly, the ability to precisely and automatically identify fake videos is highly desirable for mitigating these threats.

\begin{figure}
    \centering
    \includegraphics[width=1.0\columnwidth]{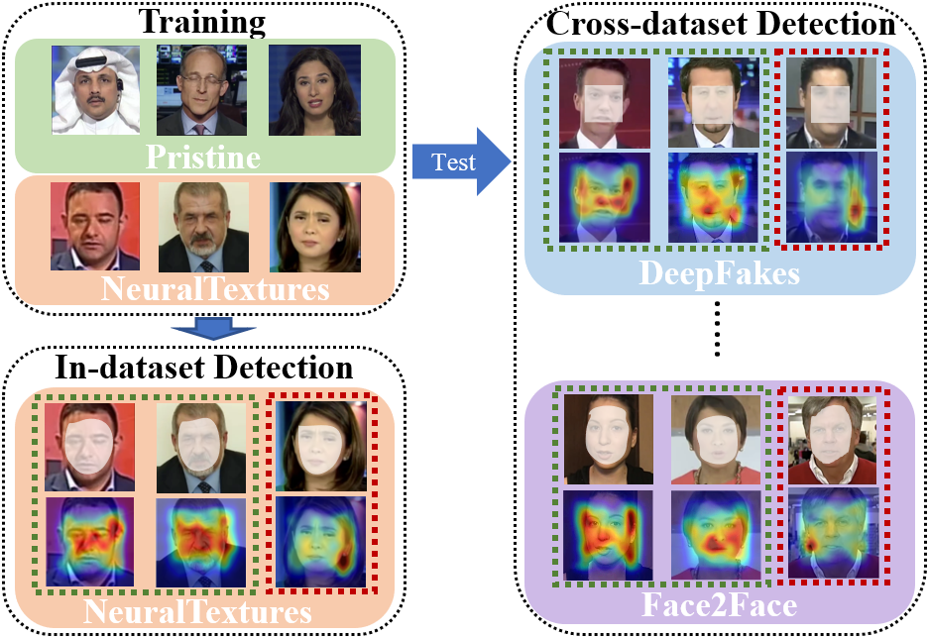}
    \caption{Salient features obtained by the Xception model trained on NeuralTextures. The {\textit{correct}} and {\textit{wrong}} predictions are respectively marked with {\textit{green}} and {\textit{red}} boxes. The shaded regions highlight the forged location annotations.}
    \label{fig1}
\end{figure}

\par At its core, deepfake detection involves identifying the subtle differences between real and synthetic images. A first class of detection methods~\cite{Face-Xray,PCL, UIA-VIT, ICT,LipForensics,DCL} leverage semantic visual clues of forgeries, such as abnormal blending boundaries~\cite{Face-Xray} and face incongruities~\cite{LipForensics}. Another line of work~\cite{F3Net, li_frequency, liu_spatial} builds upon the specific domain features, \textit{e.g.} the up-sampling artifacts~\cite{liu_spatial} in the spectrogram, which vary according to the authenticity of images. Upon scrutinizing the released implementations of existing methods, we empirically observe that current methods still suffer from two issues: (1) As deepfake techniques improves, such perceivable visual artifacts are significantly weakened, potentially compromising the reliability of deepfake detection. (2) Certain methods focus only on particular image regions, such as blending boundaries~\cite{Face-Xray,SBIs}, mouths and eyes~\cite{LipForensics,liu2022cross,dong2022towards}, and could neglect other regions where forgery clues may be abundant. Additionally, the poor generalization of these methods on unseen datasets, as demonstrated by the low performance on the Deepfake Detection Challenge preview benchmark dataset, highlights the need for further improvement in this area. The best-performing model achieved only 0.797 frame-level AUC when trained on the FaceForensics++ dataset, leaving considerable room for improvement.

\par The forgery clues on a synthesized face are typically not evenly distributed, where most regions reserve the pristine image content and forgery clues are often found in the synthesized regions~\cite{Face-Xray,ICT,two-branch, Explaining}. As such, the key to deep forgery detection lies in correctly identifying such forgery regions. Interestingly, we performed a saliency analysis of the features extracted by the Xception model~\cite{Xception}, commonly used as the backbone for deepfake detection, as visualized in Fig. {\ref{fig1}}. We find that the responses of the Xception model are universally diffuse and may encompass non-forgery or even non-facial regions for both in-dataset and cross-dataset scenarios. However, for cases of correct predictions, the model inadvertently focuses on manipulated regions when it is successful at exposing forgery. We could attribute the drawback of existing deepfake detection methods to their tendency to focus on particular visual clues and domain specificities, while failing to identify other manipulated regions, thus not being able to maximally uncover evidence for detecting forgeries.

\par
\begin{figure}[t]
    \centering
    \includegraphics[width=1.0\columnwidth]{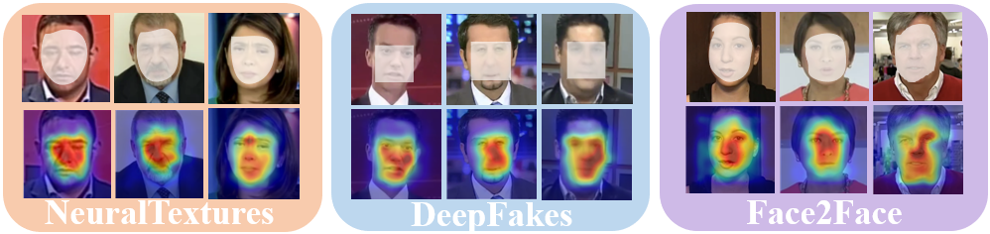}
    \caption{Salient features from our method trained on NeuralTextures. Compared to Xception model, our model can better capture forgery artifacts from the potential forged regions.}
    \label{fig2}
\end{figure}

 Building upon these insights, we propose to explicitly locate forgery regions as an intermediate objective to guide our forgery detection task. We adopt two input streams consisting of RGB images as well as Spatial Rich Model~\cite{srm} (SRM) filtered images, which are frequently used as a supplemental input to RGB images for capturing high-frequency components. Instead of directly fusing two modalities~\cite{Local-realtion,SOLA}, we propose a Cross-modality Consistency Enhancement (CMCE) module that collaboratively learns a combined representation with preserving the informative features in each modality. Subsequently, this combined feature is passed through two downstream networks, namely a localization branch that serves to detect all plausible forgery regions, as well as a classification branch that extracts forgery clues based on the detected forgery regions, facilitated by our Local Forgery Guided Attention (LFGA) module. Since forged location annotations are generally unavailable, we propose a Semi-supervised Patch Similarity Learning (SSPSL) strategy to estimate patch-level forged location annotations. We also design a Multi-scale Patch Feature Fusion (MPFF) module which allows capturing of prominent artifacts in the shallow levels of two downstream networks, while maintaining location consistency of each image patch. Examples of salient features extracted by our model are illustrated in Fig. {\ref{fig2}}, indicating that our method is able to locate important forgery regions.

\par To evaluate the effectiveness of our method, we conduct extensive experiments on six widely used benchmark datasets, \textit{i.e}., FaceForensics++~\cite{Xception}, two versions of CelebDF~\cite{cd1, cd2}, DeepFakeDetection~\cite{dfd}, Deepfake Detection Challenge preview~\cite{dfdc_p} and DeeperForensics 1.0~\cite{df1.0}. Our method performs well with or without forged location annotations and significantly outperforms previous methods with respect to generalization to unseen forgeries. To summarize, the key contributions of this work are as follows:
\begin{itemize}
\item {We propose an innovative framework for deepfake detection that effectively focuses on the potential forged regions to capture adequate evidence for forgery detection, with remarkable generalization on unseen forgeries.}
\item {We propose three functional modules in our model to take full advantage of RGB images and SRM noise residuals, by combining multi-modal features and multi-scale patch features.}
\item {We devise a semi-supervised patch similarity learning strategy to effectively supervise the detection of forgery regions even though such annotations are unavailable.}
\end{itemize}

\section{Related work}
\par Deepfake detection is often regarded as a binary classification task, where overfitting severely impacts the model's generalization performance on unseen datasets. In this context, some work~\cite{two-branch, F3Net, Luo, liu_spatial, li_frequency, qian2020, Leveraging, tan2023learning} extend facial semantic features and propose to detect forgery artifacts through high-frequency features that are challenging to identify within the texture content. Das et al.~\cite{das2021towards} present a dynamic data augmentation to alleviate the overfitting problem to significant semantic visual artifacts. Multi-Attention~\cite{Multi-attetion} formulates forgery detection as a fine-grained classification problem and proposes the multi-attention mechanism to enhance textural and semantic features. Xception-Reg~\cite{Xception_Reg} utilizes an attention mechanism to highlight informative regions, thereby improving the binary classification. Although these methods achieved good results on in-datasets, their evaluation on cross-datasets was unsatisfactory.

\par A large body of literature~\cite{Face-Xray, SBIs, ICT, SOLA, DCL, PCL, UIA-VIT, LTTD} explores semantic visual clues of forgeries to compensate for the limitations of single classification features. Face X-Ray~\cite{Face-Xray} highlights blending as a common operation in face swap and seeks to uncover evidence of blending for justification of manipulation. Chen et al. ~\cite{chen_self} explore blending-based forgeries in greater detail by analyzing facial features including the eyes, nose, and mouth, as well as considering the blending ratios. SBIs~\cite{SBIs} develops a proprietary dataset for self-blended images by employing the synthesis approach in~\cite{Face-Xray} and avoids overfitting to manipulation-specific artifacts.~\cite{Local-realtion, DCL, PCL, UIA-VIT, LTTD} determine image authenticity by identifying inconsistencies within the image and propose consistency loss for local image patches. Additionally, ~\cite{ICT, Id-reveal, Implicit} leverage identity inconsistency of the manipulated images, and SOLA~\cite{SOLA} captures forgery anomalies by enhancing the local patch differences. LipForensics~\cite{LipForensics} proposes a spatio-temporal network to learn high-level semantic irregularities in mouth movements. Nevertheless, complex synthesis and post-processing methods may weaken these artifacts while focusing on a few specific forgery regions ignores other possible forgery cues, thus reducing their applicability to particular datasets.

\par Our method differs from the methods mentioned above as we do not limit ourselves to such fixed artifacts that are mostly due to inherent defects in earlier deepfake generation algorithms - such as blending boundaries, patch inconsistencies, face incongruities, upsampling artifacts, etc. We intend to explicitly identify potential forgery regions in manipulated images to facilitate the extraction of forgery evidence while minimizing interference from non-forgery regions (such as the duplicated background).

\begin{figure*}[t]
    \centering
    \includegraphics[width=0.975\textwidth, height=0.36\textheight]{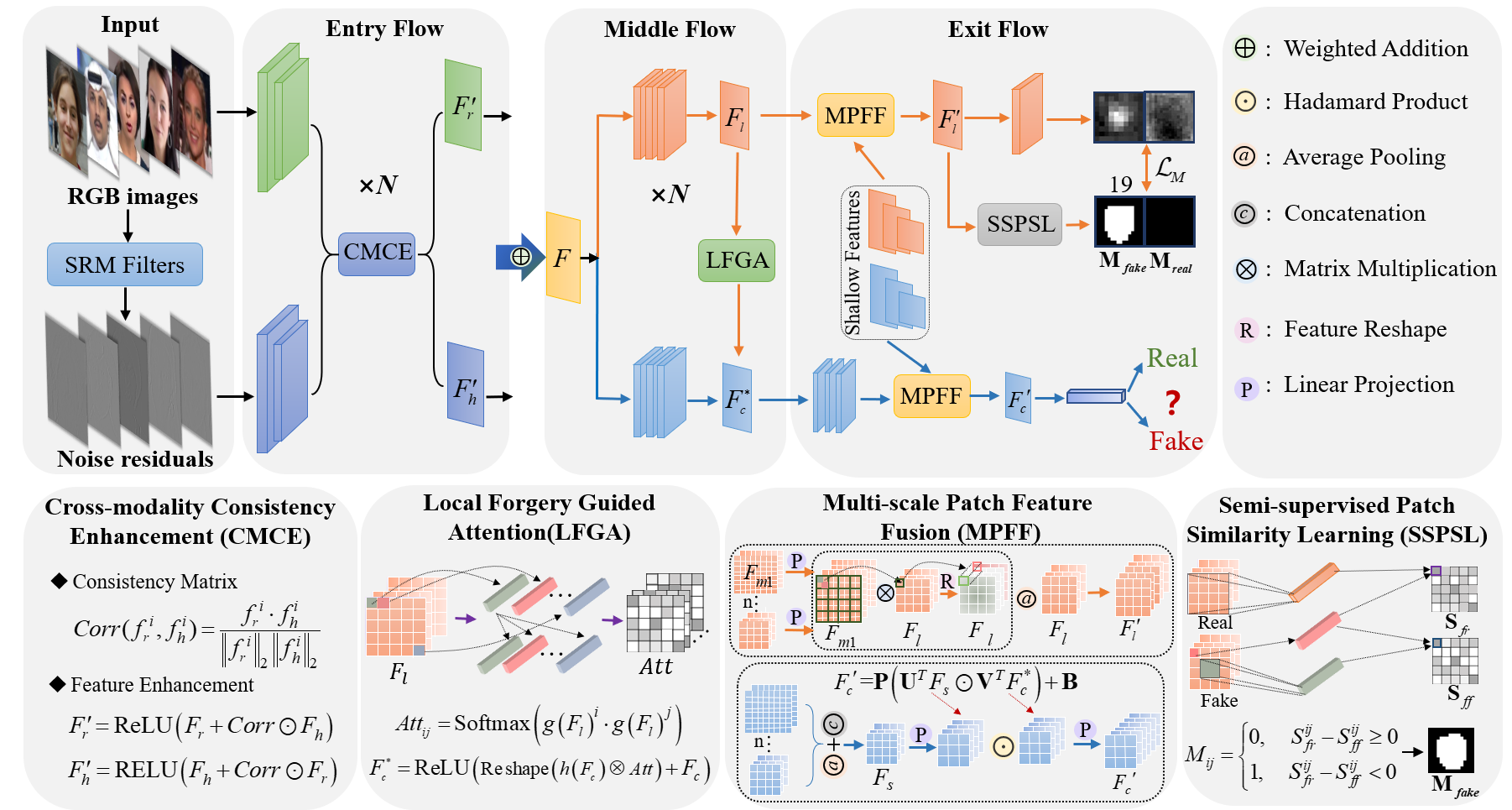}
    \caption{Overview of our framework. In the entry flow, we employ the CMCE module $N=3$ times to collaboratively learn features $F$ from two modalities. In the middle flow, we obtain location features $F_l$ and classification features $F_c$. $F_l$ is supervised for regional forgery detection and also serves to boost the classification features by our LFGA modules. In the exit flow, we design MPFF modules to incorporate multi-scale information for each branch while maintaining patch location consistency. Finally, we introduce a semi-supervised strategy (SSPSL) for training the localization branch without fine forgery annotations.}
    \label{fig3}
\end{figure*}

\section{Proposed method}
We tackle the problem of building a generalizable face forgery classifier by effectively identifying all potential forged regions for uncovering sufficient forgery artifacts. Generally, different parts of one synthesized face would have an uneven distribution of forgery artifacts produced by face manipulation techniques, where some regions contain ample forgery clues while others are not manipulated, retaining the pristine image content. Naturally, effectively locating potential manipulating regions would be greatly useful for forgery detection. As such, we propose to explicitly model the detection of such regions, where we design a two-stream framework comprising a localization branch and a classification branch. The localization branch determines whether an image patch contains forgeries, which provides an attention-weighted guidance for the classification branch to focus on more probable forgery patches for uncovering forgery clues. 

An overview of our proposed framework is illustrated in Fig. \ref{fig3}. Firstly, we adopt a dual stream input, consisting of the RGB image and its associated Spatial Rich Model (SRM)~\cite{srm} noise residuals, since SRM better captures high-frequency features crucial for image forensics. We combine the two modalities through our Cross-Modality Consistency Enhancement (CMCE) module to obtain a combined feature representation. Subsequently, this feature passes through two downstream networks, namely a localization branch that detects possible forgery in image patches, and a classification branch that extracts forgery clues for determining whether the image has been manipulated. We propose a Local Forgery Guided Attention (LFGA) module which derives attention maps from the location branch to enhance the extraction of classification features. To enhance information retention at multi-scales, we also designed Multi-scale Patch Feature Fusion (MPFF) modules for each branch. Furthermore, due to the shortage of annotations for forgery locations, we also enact a Semi-supervised Patch Similarity Learning (SSPSL) strategy. In what follows, we will discuss the details of each key component in our approach.

\subsection{Cross-modality Consistency Enhancement}
\par Our CMCE module performs collaborative learning to learn a combined representation from RGB and SRM modalities. Different from previous methods~\cite{SOLA, two-branch, Local-realtion}, we refrain from merging two modalities through direct concatenation or attention-weighted enhancement. Instead, we seek to ensure that both branches preserve their respective characteristics as much as possible, while also capturing the interaction and interplay between the two modalities.

\par Specifically, the inputs to the CMCE module consist of the RGB modality feature map ${F_r} \in {\mathbb{R}^{c \times h \times w}}$ and the SRM modality feature map ${F_h} \in {\mathbb{R}^{c \times h \times w}}$. We compute a cross-modal consistency map via an element-wise inner product
\begin{equation}
    Corr({f_r^i},f_h^i) = \frac{{{f_r^i} \cdot f_h^i}}{{||{f_r^i}|{|_2}||f_h^i|{|_2}}},
    \label{eq1}
\end{equation}
where ${f_r^i} \in {\mathbb{R}^{c \times 1 \times 1}}$, ${f_h^i} \in {\mathbb{R}^{c \times 1 \times 1}}$, and $i \in \left\{ {1,...,hw} \right\}$. Subsequently, we apply the correlation map $Corr$ to $F_r$ and $F_h$ through:
\begin{equation}
    \begin{array}{l}
        F_r' = \text{ReLU}\left( {F_r + Corr \odot {F_h}} \right),\\
        {{F'_h}} = \text{ReLU}\left( {{F_h} + Corr \odot F_r} \right).
    \end{array}
    \label{eq2}
\end{equation}
We repeat the above for $N=3$ times for thorough cross-modal learning and we sum the two feature maps to obtain:
\begin{equation}
    \begin{array}{l}
        F = {{F'_r}} + {{F'_h}}.
    \end{array}
\end{equation}
\par
\begin{figure}[t]
    \centering
    \setlength{\abovecaptionskip}{-0.01cm}
    \includegraphics[width=0.9\columnwidth]{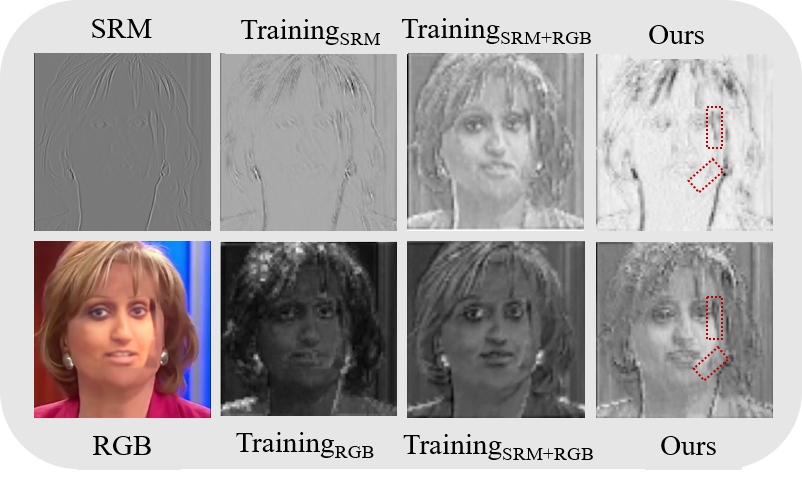}
    \caption{Example visualization of features obtained by our method and baselines. Training${\rm{_{_{RGB}}}}$ and Training${\rm{_{_{SRM}}}}$ denote features of the Xception model trained on a single modality. Training${\rm{_{_{RGB+SRM}}}}$ are obtained by a two-stream model with the direct summation of two modalities.}
    \label{fig4}
\end{figure}

 Fig. \ref{fig4} demonstrates the original images, the features of single-modal input (Training${\rm{_{_{SRM}}}}$ and Training${\rm{_{_{RGB}}}}$), the features with the direct summation two modalities (Training${\rm{_{_{SRM+RGB}}}}$) and the features of CMCE. From the figure, we may observe that: (1) The CMCE module learn richer forgery features compared with the single modality. (2) It also preserve independent and representative features for two modalities. The differences are more obvious in the SRM modality, especially the direct summation of two modalities will lead to SRM and RGB features being very similar.

\subsection{Local Forgery Guided Attention}
\par The key to assessing if an image has been tampered with lies in effectively garnering evidence. As discussed in the introduction, a common failure case in existing methods occurs when these models heavily rely on non-manipulated image regions to make their predictions. As such, we believe a potent approach to tackling this issue would be to train our model in identifying any manipulated regions with greater confidence, which would serve to better extract forensic evidence. To achieve this, we explicitly include a localization branch for locating probable forgery regions. We employ our Local Forgery Guided Attention (LFGA) module to obtain an attention map from the location features to guide the learning of more robust and informative classification features. 

Specifically, we denote the intermediate feature map from the localization branch as ${F_l}\in {\mathbb{R}^{\tilde c \times \tilde h \times \tilde w}}$ and that from the classification branch as ${F_c}\in {\mathbb{R}^{\tilde c \times \tilde h \times \tilde w}}$. We first learn self-attention maps $Att \in {\mathbb{R}^{{\tilde h\tilde w} \times {\tilde h \tilde w}}}$ for ${F_l}$ via:
\begin{equation}
    {Att_{ij}} = \text{Softmax}\left( {g\left( {F_l} \right)^i \cdot g\left( {F_l} \right)^j} \right),
    \label{eq4}
\end{equation}
where $g$ denotes a linear transformation and $i, j \in \left\{ {1,...,\tilde h\tilde w} \right\}$ are the indices. The self-attention maps $Att$ identify image patches with similar characteristics and correspond to saliency representations for forgery likelihood. We then apply a transformation $h$ on the classification feature map $F_c$, followed by a matrix multiplication with the attention maps $Att$ to enhance the classification feature with more location-aware information:
\begin{equation}
    {F_c}^*  = \text{ReLU} \left( {\text{Reshape}\left(h\left( {{F_c}} \right) \otimes Att \right) + {F_c}} \right).
    \label{eq5}
\end{equation}
\par We also apply the LFGA module for $N=3$ times. This allows multiple scale learning of the location-enhanced classification feature ${F_c}^*$.

\subsection{Multi-scale Patch Feature Fusion}
\par Many existing works for deepfake detection fail to take advantage of the fact that the artifacts resulting from forgery methods may be more dominant in the shallow features. For instance, as illustrated in Fig. \ref{fig4}, the traces created by image blending are visually prominent. One strategy to uncover such artifacts in a robust fashion would be to examine them at multiple scales for two branches. Overall, the classification features concern global semantic information and localization features focus on local spatial details. Moreover, it is necessary to maintain location information for each image patch, as it plays a crucial role in our model. To this end, we design two Multi-scale Patch Feature Fusion (MPFF) modules.

\par For the localization branch, we denote the last layer feature as ${F_l} \in {\mathbb{R}^{c_1 \times h_1 \times w_1}}$ and a intermediate feature map as ${F_{ml}} \in {\mathbb{R}^{c_2 \times h_2 \times w_2}}$ $\left( {{h_2} > {h_1},{w_2} > {w_1}} \right)$. Due to the expansion of receptive field after several layers, ${F_l}$ may lose their discrimination in representing local regions. So we spatially divide ${F_{ml}}$ into $h_1 \times w_1$ non-overlapping patches ${P_k}$ with necessary zero-padding, where ${k = \{ 1,...,h_1w_1\} }$. Then we calculate intra-patch consistency map ${F_{ml}^\prime} \in {\mathbb{R}^{h_2 \times w_2}}$ for the different scale features ${F_l}$ and ${F_ml}$ by:
\begin{equation}
    {{f_{ml}^{(k,j)}}}^\prime = \text{Tanh} \left( {\frac{{\theta (p_k^j) \cdot \theta (f_l^k)}}{c}} \right),
    \label{eq6}
\end{equation}
where $p_k^j$ is the $j$th feature vector of $P_k$, $f_l^k$ is the $k$th feature vector of $F_l$, $\theta$ is an embedding function realized by $1 \times 1$ convolutions, and $c$ is the embedding dimension. ${F_{ml}^\prime}$ is finally reshaped to the same scale $(h_1, w_1$) with $F_l$. We perform the operation to all intermediate features and concatenate them with $F_l$ to obtain a final multi-scale localization feature $F_l^{\prime}$. Finally, we pass $F_l^{\prime}$ into a prediction head consisting of a single 1x1 convolution layer.

\par To retain the spatial relationships among image patches across different scales, we adopt the Low-rank Bilinear Pooling~\cite{bilinear} to consolidate classification features. The feature $F_c^*$ from the classification stream is processed by a convolutional block, and all shallow features are resized into the same scale as $F_c^* \in {\mathbb{R}^{c^* \times h^* \times w^*}}$ via average pooling and concatenated together, denoted as $F_s \in {\mathbb{R}^{c_s \times h^* \times w^*}}$. We obtain a final classification feature as:
\begin{equation}
    {F_c^\prime}{\rm{ = }}{{\bf{P}}}\left( {{\bf{U^T}}{F_{s}} \odot {\bf{V^T}}{F_c^*}} \right) + \bf{{B}},
    \label{eq7}
\end{equation}
where ${\bf{P}} \in {\mathbb{R}^{n \times m}},{\bf{U}} \in {\mathbb{R}^{c_s \times m}},{\bf{V}} \in {\mathbb{R}^{c^* \times m}}$ are learned projection matrices, and ${\bf{B}} \in {\mathbb{R}^{n \times h^* \times w^*}}$ is a bias map. $F_c^\prime$ is fed into a standard classification head to predict the final results.

\subsection{Semi-supervised Patch Similarity Learning}
\par Since the majority of public deepfake datasets do not include annotations for forgery locations, we devise a Semi-supervised Patch Similarity Learning (SSPSL) strategy to train our localization branch, drawing inspirations from ~\cite{hjelm_earning, dong2023region}.

\begin{figure}[t]
    \centering
    \setlength{\abovecaptionskip}{-0.02cm}
    \includegraphics[width=1.0\columnwidth]{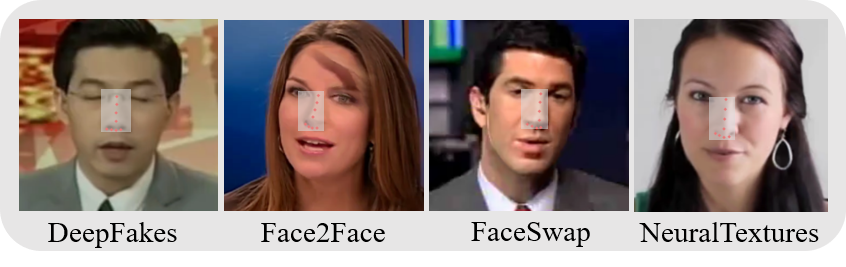}
    \caption{The positions of noses and the designated forgery regions for some samples in the FF++ dataset.}
    \label{fig5}
\end{figure}

\par Forgery location maps for real images are always fixed as all zeroes. For fake images, we may not have access to forgery annotations, but our analysis can ascertain that specific facial regions - such as the nose, eye, and mouth - have been manipulated, and they are also considered sensitive regions for forgery detection. Consequently, we can approximately select features for sensitive facial patches to represent the manipulated face region's distribution.

\par Specifically, we first utilize facial landmarks to detect nose positions of the fake image and designate a rectangular region as the manipulated region, as shown in Fig. \ref{fig5}. We treat all real images within a batch as positive samples and all manipulated regions of fake images within a batch as negative samples. We denote by ${f_r \in {\mathbb{R}^{c \times 1 \times 1}}}$ the average anchor from real samples, ${f_a \in {\mathbb{R}^{c \times 1 \times 1}}}$ the average anchor from fake samples, and ${F_f}$ the feature maps from fake samples. Then we obtain a similarity map ${\bf{S}}_{fr}$ between ${F_f}$ and $f_r$ via an element-wise inner product: 
\begin{equation}
    {S_{fr}^{ij}} = \frac{{{f_f^{ij}} \cdot f_r}}{{||{f_f^{ij}}|{|_2}||f_r|{|_2}}},
    \label{eq8}
\end{equation}
where $(i,j)$ indexes the the spatial position in $F_f$. In the identical fashion, we obtain a similarity map ${\bf{S}}_{ff}$ by performing the same operation on each local-global pair of $f_f^{ij}$ and $f_a$. 

Consequently, for a fake image, we define the predicted location annotation ${\bf{M}} \in {\mathbb{R}^{h_1 \times w_1}}$ as a binary comparison map. When $S_f^{ij}$ is in close proximity to $f_r$, the patch is predicted as not containing forgeries; otherwise, it is deemed to contain forgeries. The process is formalized as:
\begin{equation}
    {M_{ij}} = \left\{ {\begin{array}{*{20}{c}}
    {0, \ \ \ \ {\rm{   }}{S_{fr}^{ij}} - {S_{ff}^{ij}} \ge  0}\\
    {1, \ \ \ \ {\rm{   }}{S_{fr}^{ij}} - {S_{ff}^{ij}} <  0}
\end{array}} \right.
    \label{eq9}
\end{equation}

\subsection{Loss functions}
\par For the classification stream, we use cross-entropy loss to supervise the final predicted probability $\hat y$ with binary labels of 0 and 1:
\begin{equation}
    {{\mathcal L}_C} =  - \left[ {y\log \hat y + \left( {1 - y} \right)\log \hat y} \right]
    \label{eq10}
\end{equation}
where ${y}$ is a binary label indicating whether the input image has been manipulated or not.

\par For the localization stream, the annotations can be estimated via the proposed SSPSL described in the preceding subsection or determined through the pixel-level annotations ${{\bf{M}}}$ given by the original datasets. For the latter, we divide ${{\bf{M}}}$ into $h_1 \times w_1$ non-overlapping patches, and then the corresponding label ${\bf{M}}_k$ $\left( {k = 1,2,...,{h_1}{w_1}} \right)$ for each patch $P_k$ is obtained by averaging all the values of ${\bf{M}}_{P_k}$:
\begin{equation}
    {{M}_k} = \left\{ {\begin{array}{*{20}{c}}
    {0, \ \ \ \ {\rm{  }}{\rm{avg}}\left( {{\bf{M}}_{P_k}} \right) = 0}\\
    {1, \ \ \ \ {\rm{  }}{\rm{avg}}\left( {{\bf{M}}_{P_k}} \right) > 0}
    \end{array}} \right..
    \label{eq11}
\end{equation}
Assume ${\bf{\hat M}}$ is the predicted location map, the accuracy of the prediction is measured by the cross entropy loss:
\begin{equation}
    {{\mathcal L}_M} = \sum\limits_{k = 0}^{{h_1}{w_1}} { - \left[ {{M_k}\log {{\hat M}_k} + \left( {1 - {M_k}} \right)\log \left( {1 - {{\hat M}_k}} \right)} \right]} 
    \label{eq12}
\end{equation}
And finally, we can train our model in an end-to-end manner with the total of two losses, described as:
\begin{equation}
    {{\mathcal L}_{total}} = {{\mathcal L}_C} + {{\mathcal L}_M}
    \label{eq13}
\end{equation}

\section{Experiments}
\subsection{Settings}
\textbf{Datasets.} We evaluate our model on six common forgery datasets: Faceforencis++~\cite{Xception} (\textbf{FF++}), two versions of Celeb-DF~\cite{cd1, cd2} (\textbf{CD1} and \textbf{CD2}), DeepFake Detection Challenge Preview~\cite{dfdc_p} (\textbf{DFDC$\_$P}), DeepFakeDetection~\cite{dfd} (\textbf{DFD}) and DeeperForensics-1.0~\cite{df1.0} (\textbf{DFo}). Our experiments utilize the high-quality version (c23) of FF++ that contains 4,000 forgery videos produced by four algorithms: DeepFakes~\cite{df} (\textbf{DF}), Face2Face~\cite{face2face} (\textbf{F2F}), FaceSwap~\cite{fs} (\textbf{FS}) and NeuralTextures~\cite{neuraltextures} (\textbf{NT}). More details are given in the appendix.

\noindent \textbf{Implementation Details.} In data pre-processing, we align official annotations of the FF++ dataset with the original videos and extract face crops. All faces in our experiments are cropped to $299 \times 299$ and uniformly normalized to $\left[ {0,1} \right]$. We utilized some common augmentations, such as flip, contrast and blur. Additionally, we used random cropping to increase the diversity of forged regions when ensuring alignment of annotations with images.
\par For training, we adopt the backbone Xception~\cite{Xception_} initialized with pretrained weights and use the Adam~\cite{adam} optimizer with betas 0.9 and 0.999, and epsilon 1e-8. The initial learning rate is set as $5{e^{ - 4}}$ and decays by $50\% $ per five epochs. The size of the forgery map predicted by the location stream is set to 19 x 19. The hyperparameters $m$ and $n$ in the MPFF module is set as $2048$ and $4096$. All experiments are implemented with PyTorch on the platform with NVIDIA RTX 3090 24GB.

\subsection{Evaluations}

\begin{table}[t]
\centering
\tabcolsep=4.4pt
\caption{In-dataset evaluation results on FF++ (AUC).}
\begin{tabular*}{\linewidth}{c|c|ccccc}
        \cline{1-7}
        Method & \ FF++ & DF &F2F & FS & NT & Avg \\
        \cline{1-7}
         Xception~\cite{Xception} & 0.963 & 0.994 & 0.995 &  0.994 &  0.995 & 0.994 \\
         Face X-Ray~\cite{Face-Xray} & 0.985 & 0.991 & 0.993 & 0.992 & 0.993 & 0.992 \\
         DCL~\cite{DCL} & 0.993 & 1.00 & 0.992 & 0.999 & 0.990 & 0.995 \\
         PCL+l2G~\cite{PCL} & 0.991 & 1.00 & 0.990 & 0.999 & 0.976 & 0.991 \\
         SOLA~\cite{SOLA} & 0.992 & 1.00 & 0.995 & 1.00 & \textbf{0.998} & 0.998 \\
         SBIs~\cite{SBIs} & 0.992 & 1.00 & 0.999 & 0.999 & 0.988 & 0.996 \\
         \cline{1-7}
         Ours & \textbf{0.998} & \textbf{1.00} & \textbf{0.999} & \textbf{1.00} & 0.994 & \textbf{0.998} \\
         Ours-$semi$ & 0.997 & 1.00 & 0.997 & 0.999 & 0.992 & 0.997\\ 
        \cline{1-7}
    \end{tabular*}\par\smallskip
\label{tab2}
\end{table} 

\begin{table}[t]
\centering
\setlength{\belowcaptionskip}{-0.1cm}
\tabcolsep=2.3pt
\caption{Benchmark results on four sub-datasets (AUC).}
\begin{tabular*}{\linewidth}{ccccccc}
        \hline 
        Training set & Method & DF & F2F & FS & NT & Avg \\[1pt] \hline 
        \multirow{6}{*}{DF} 
            & Xception~\cite{Xception} & \cellcolor{gray!20} 0.993 & 0.736 & 0.490 & 0.736 & 0.739 \\
            & Face X-Ray~\cite{Face-Xray} & \cellcolor{gray!20} 0.987 & 0.633 & 0.600 & 0.698 & 0.730 \\
            & DCL~\cite{DCL} & \cellcolor{gray!20} 1.00 & 0.771 & 0.610 & 0.750 & 0.782 \\
            & Ours & \cellcolor{gray!20} \textbf{1.00} & \textbf{0.864} & \textbf{0.659} & \textbf{0.842} & \textbf{0.841}\\
            & Ours-$semi$ & \cellcolor{gray!20} 0.999 & 0.831 & 0.571 & 0.823 & 0.806 \\
        \hline 
        \multirow{6}{*}{F2F} 
            & Xception~\cite{Xception} & 0.803 & \cellcolor{gray!20} 0.994 & 0.762 & 0.696 & 0.814 \\
            & Face X-Ray~\cite{Face-Xray} & 0.630 & \cellcolor{gray!20}0.984 & \textbf{0.938} & \textbf{0.945} & 0.874 \\
            & DCL~\cite{DCL} & \textbf{0.919} & \cellcolor{gray!20}0.992 & 0.596 & 0.667 & 0.794 \\
            & Ours & 0.826 & \cellcolor{gray!20} \textbf{0.999} & 0.901 & 0.905 & \textbf{0.908}\\
            & Ours-$semi$ & 0.810  & \cellcolor{gray!20} 0.997 & 0.882 & 0.891 & 0.895\\
        \hline 
        \multirow{6}{*}{FS} 
            & Xception~\cite{Xception} & 0.664 & 0.888 & \cellcolor{gray!20} 0.994 & 0.713 & 0.815 \\
            & Face X-Ray~\cite{Face-Xray} & 0.458 & \textbf{0.961} & \cellcolor{gray!20} 0.981 & \textbf{0.957} & 0.839\\
            & DCL~\cite{DCL} & \textbf{0.748} & 0.698 & \cellcolor{gray!20} 0.999 & 0.526 & 0.743\\
            & Ours & 0.706 & 0.933 & \cellcolor{gray!20} \textbf{1.00} & 0.905 & \textbf{0.886} \\
            & Ours-$semi$ & 0.679 & 0.927 & \cellcolor{gray!20} 0.999 & 0.885 & 0.872 \\
        \hline 
        \multirow{6}{*}{NT} 
            & Xception~\cite{Xception} & 0.799 & 0.813 & 0.731 & \cellcolor{gray!20} 0.991 & 0.834 \\
            & Face X-Ray~\cite{Face-Xray} & 0.705 & 0.917 & 0.910 & \cellcolor{gray!20} 0.989 & 0.880 \\
            & DCL~\cite{DCL} & \textbf{0.912} & 0.521 & 0.783 & \cellcolor{gray!20} 0.990 & 0.802 \\
            & Ours & 0.869 & \textbf{0.969} & \textbf{0.946} & \cellcolor{gray!20} \textbf{0.994} & \textbf{0.945} \\ 
            & Ours-$semi$ & 0.836 & 0.954 & 0.935 & \cellcolor{gray!20} 0.991 & 0.929 \\
        \hline 
    \end{tabular*}\par\smallskip
\label{tab3}
\end{table}

\noindent \textbf{In-dataset performance.} In in-dataset evaluations, we benchmark our model against state-of-the-art methods on FF++. The results are shown in  Tab. \ref{tab2}, where Ours-$semi$ indicates that during the training process, the SSPSL module is used to estimate the forgery annotations. From Tab. \ref{tab2}, we observe that the results of our method are superior to previous methods that have already achieved remarkable performance. Specifically, our method surpasses the best competitor DCL by $0.5 \%$ in terms of AUC. Ours-$semi$ obtains similar results to the supervised one, partially confirming the effectiveness of the SSPSL module.

\noindent \textbf{Cross-dataset performance.} Evaluating generalization performance in a cross-dataset setting is crucial for real-world applications because images may originate from unknown or uncertain forgery methods. Despite existing methods achieving good results on the in-dataset setting, their robustness and generalizability remain a major shortcoming when applied to cross-dataset detection.

\begin{table*}[t]
\centering
\caption{The results of our model trained on FF++ and evaluated on the other benchmarks. The last row $\left( * \right)$ complements ACC. The abbreviation 'PrD' stands for private data. Bold and blue fonts are used to indicate the best and second-best performances. }
\begin{tabular*}{\linewidth}{ccccc|cccccc}
        \hline 
        \multicolumn{5}{c|}{Frame-level} & \multicolumn{6}{c}{Video-level} \\ \hline 
        Method & Training set & CD2 & DFDC$\_$P & DFD & Method & Training set & CD1 & CD2 & DFDC$\_$P & DFo \\ \hline 
         Xception~\cite{Xception} & \ FF++ & 0.655 & 0.722 & 0.705 & Xception~\cite{Xception} & \ FF++ & 0.623 & 0.737 & -- & 0.845 \\
         Face X-Ray~\cite{Face-Xray} & \ FF++ & 0.7520 & 0.700 & 0.935 & Face X-Ray~\cite{Face-Xray} & PrD & 0.806 & -- & -- & 0.868 \\
         Luo.\textit{et al.}~\cite{Luo} & \ FF++ & 0.794 & {\color{blue}0.797} & 0.919 & FWA~\cite{FWA} & PrD & 0.538 & 0.569 & -- & 0.502 \\
         Multi-Attention~\cite{Multi-attetion} & \ FF++ & 0.674 & 0.663 & 0.755 & DAM~\cite{DAM} & \ FF++ & -- & 0.783 & 0.741 & -- \\
         LTW~\cite{LTW} & \ FF++ & 0.771 & 0.746 & 0.886 & Li.\textit{et.al}~\cite{li_wavelet} & \ FF++ & -- & 0.870 & 0.785 & --\\
         PCL+l2G~\cite{PCL} & PrD & 0.818 & 0.744 & -- & FTCN~\cite{FTCN} & \ FF++ & -- & 0.869 & 0.740 & -- \\
         Local-relation~\cite{Local-realtion} & \ FF++ & 0.783 & 0.765 & 0.892 & LiSiam~\cite{LiSiam} & \ FF++ & {\color{blue}0.811} & 0.782 & -- & -- \\
         DCL~\cite{DCL} & \ FF++ & 0.823 & -- & 0.917 & SBIs~\cite{SBIs} & PrD & -- & 0.870 & {\color{blue}0.822} & -- \\
         ICT~\cite{ICT} & PrD & {\color{blue}0.857} & -- & 0.841 & LipForensics~\cite{LipForensics} & \ FF++ & -- & 0.824 & -- & 0.976 \\
         UIA-ViT~\cite{UIA-VIT} & \ FF++ & 0.824 & 0.758 & {\color{blue}{0.947}} & LTTD~\cite{LTTD} & \ FF++ & -- & {\color{blue}0.893} & -- & {\color{blue}0.985} \\ \hline
         Ours & \ FF++ & \textbf{0.860} & \textbf{0.835} & \textbf{0.955} & Ours & \ FF++ & \textbf{0.847} & \textbf{0.922} & \textbf{0.897} & \textbf{0.990}\\ \hline 
         \hline
         Ours$^*$ (ACC $\%$) & \ FF++ & 78.17 & 71.25 & 88.19 & Ours$^*$ (ACC $\%$) & \ FF++ & 75.00 & 84.60 & 75.70 & 90.54\\ \hline 
    \end{tabular*}\par\smallskip
\label{tab4}
\end{table*}

\begin{table}[t]
\centering
\tabcolsep=5pt
\caption{The performance (AUC) of two training strategies.}
\begin{tabular*}{\linewidth}{c|cccc|cc}
        \hline
        & \multicolumn{4}{c|}{Frame-level} & \multicolumn{2}{c}{Video-level} \\ \hline
        Method & CD1 & CD2 & DFDC$\_$P & DFD & CD1 & CD2 \\
        \hline
         Ours & \textbf{0.803} & \textbf{0.860} & \textbf{0.835} & \textbf{0.955} & \textbf{0.847} & \textbf{0.922} \\ [1pt]
         Ours-$semi$ & 0.791  & 0.837 & 0.819 & 0.947 & 0.830 & 0.893\\
        \hline
    \end{tabular*}\par\smallskip
\label{tab5}
\end{table}

\begin{figure*}
    \centering
    \includegraphics[scale=0.74]{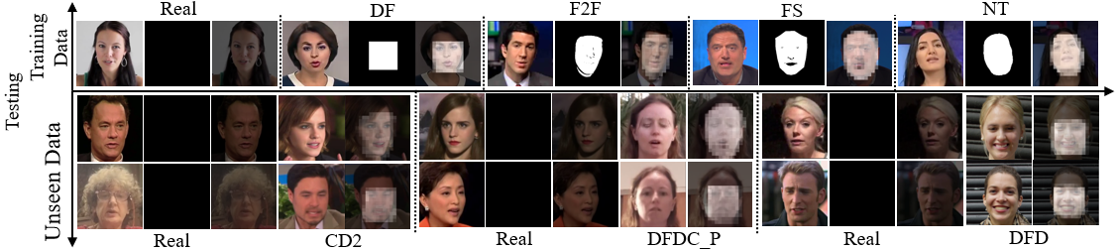}
    \caption{Predicted forgery regions of training datasets and unseen datasets from our model, trained on the FF++ dataset. }
    \label{fig6}
\end{figure*}

\par Firstly, we evaluate our model on DF, F2F, FS and NT, with the results shown in Tab. \ref{tab3}. Our methods, including the weaker model with the SSPSL strategy, outperform the competitors in most cases, particularly regarding the average AUC. For instance, our model demonstrates significant improvements of over $5\%$ in the average AUC of four sub-datasets.

\par We also evaluate our model with state-of-the-art deepfake detection methods. Our model are trained on FF++ and tested on unseen datasets, including CD1, CD2, DFD, DFDC$\_$P and DFo. The experimental results in terms of frame-level and video-level AUC are demonstrated in Tab. \ref{tab4}. We have two observations:(1) the performance of the existing deepfake detection methods on the unseen dataset is still unsatisfactory; (2) our method outperforms the best competitions for both frame-level and video-level evaluations. For example, our method improves the frame-level AUC on CD2 from 0.857 (ICT~\cite{ICT}) to 0.860, on DFDC$\_$P from 0.799 (Luo.\textit{et.al}~\cite{Luo}) to 0835. We also report the classification accuracy (ACC) of our method to provide a comprehensive detection assessment. 

\par We further evaluate the generalization performance of the SSPSL strategy on cross-datasets, and the results are presented in Tab. \ref{tab5}. As can be seen, the absence of ready-made forgery annotations leads to a performance degradation of our semi-supervised method, but the results continues to outperform state-of-the-art methods on CD1, DFDC$\_$P, and DFD. For instance, we achieve a $1.9\%$ improvement over LiSiam~\cite{LiSiam} (0.811) on CD1, and a $2.2\%$ improvement over Luo.\textit{et. al}~\cite{Luo} (0.797) on DFDC$\_$P. Our semi-supervised method, obtains a lower frame-level AUC compared to ICT~\cite{ICT} (0.857) on CD2, but ICT was trained on a private dataset created by a simulated forgery method, whereas we employ only the standard training set from the FF++ dataset. 

\par The results show that while the state-of-the-art methods may generalize well to specific datasets in a cross-dataset setting, they are unable to consistently deliver across all these datasets. In contrast, our method consistently demonstrates an overall improvement across all five cross-domain datasets, which provides strong evidence of its generalizability. By explicitly pinpointing and directing attention towards manipulated regions of fake images, our method facilitates the identification of sufficient forgery evidence while minimizing interference from non-forgery regions, thus mitigating the overfitting of the model. 

\begin{figure*}
    \centering
    \includegraphics[scale=0.53]{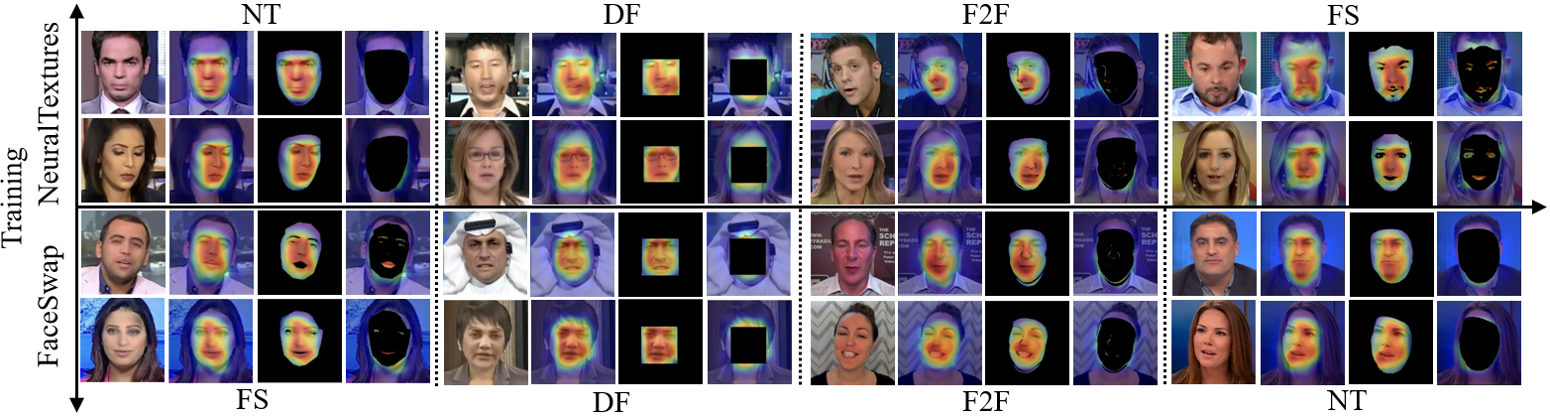}
    \caption{Grad-CAM maps from the classification stream of our model. The models are trained on two sub-datasets (NT and FS) and tested on other sub-datasets. We provide the corresponding image masks for comparative analysis.}
    \label{fig7}
\end{figure*}

\begin{figure*}
    \centering
    \setlength{\belowcaptionskip}{0.1cm}
    \includegraphics[scale=0.75]{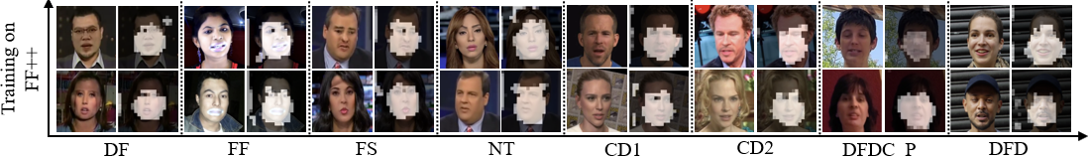}
    \caption{Estimated forgery annotations for our model with the SSPSL strategy. The model is trained on FF++.}
    \label{fig8}
\end{figure*}

\noindent \textbf{Visualization.} Our investigation also extends to the interpretability of our method. We initially conduct an analysis of our location branch and Fig .\ref{fig6} shows the predicted location maps for different datasets. The model is trained on FF++ and evaluated on both training and unseen datasets (CD2, DFDC$\_$P and DFD). Note that we have utilized intra-patch padding to ensure that the size of location maps matches that of the images. The results provide a meaningful observation that the location branch can effectively forecast the manipulated regions of fake images, even for unfamiliar data. This is significant for our model, as it ensures to guide our classification branch towards such important regions with even greater precision.

\par We further visualize classification features of our model, which is trained on two sub-datasets (FS and NT) and subsequently evaluated on the remaining three sub-datasets. The Gradient-weighted Class Activation Mapping~\cite{cam-grad} (Grad-CAM) maps are shown in Fig. \ref{fig7}. We observe that for both cross-domain and intra-domain evaluations, our method effectively and precisely focuses on the manipulated regions of an image that contain significant forgery traces while disregarding the image background, which indicates thses importance regions are greatly considered by the final classifier.

\par In Fig. \ref{fig8}, we present forgery annotations predicted by our model with the SSPSL strategy. The model is evaluated on four training sub-datasets as well as four unseen datasets. Our observations are as follows: (1) The SSPSL module effectively differentiates patch embeddings between the original background and the manipulated regions. (2) The predicted annotations proficiently highlight the manipulated regions of fake faces. This further illustrates the effectiveness of SSPSL in training our localization branch.

\begin{table}[t]
\centering
\tabcolsep=0.11cm
\caption{Ablation study of different model components.}
\begin{tabular*}{\linewidth}{cccccc}
    \cline{1-6}
    \multirow{2}{*}{Method} & \multirow{2}{*}{Training set} & \multicolumn{2}{c}{DFD}       & \multicolumn{2}{c}{CD2}       \\ \cline{3-6}
    &                   & \multicolumn{1}{c}{ACC} & AUC & \multicolumn{1}{c}{ACC} & AUC \\  \cline{1-6} 
    RGB & \multirow{4}{*}{FF++} & 83.89 & 0.911 &  67.57 & 0.801 \\ 
    SRM &                   & 84.29 & 0.925 & 71.20 & 0.821 \\ [1pt]
    Two-stream &                   & 86.29 & 0.942 & 73.28 &  0.836   \\
    Two-att-stream &  & 84.72 & 0.932 &  73.03 & 0.828 \\ [1pt]
    \cline{1-6}
    CMCE & \multirow{3}{*}{FF++} & 86.40 & 0.945 & 74.47 & 0.848 \\ 
    CMCE+LFGA &                   & 87.95 &  0.951 & 76.84 & 0.856\\ 
    CMCE+LFGA+MPFF &                   & \textbf{89.19} & \textbf{0.955} & \textbf{78.17} & \textbf{0.860} \\
    \cline{1-6}
    \end{tabular*}\par\smallskip
\label{tab6}
\end{table}

\begin{table}[t]
\setlength{\abovecaptionskip}{0.4cm}
\centering
\tabcolsep=10pt
\caption{Cross-dataset AUC of different reference regions.}
\begin{tabular*}{\linewidth}{c|cccc}
        \cline{1-5}
        Regions & Nose & Mouth & Eyes & Inner face \\ 
        \cline{1-5}
        CD2 & 0.837 & 0.827 & 0.823 & 0.840 \\
         DFDC$\_$pre & 0.819 & 0.821 & 0.816 & 0.826\\
        \cline{1-5}
    \end{tabular*}\par\smallskip
\label{tab7}
\end{table}

\subsection{Ablation Study}

\par We first assess the efficacy of each module in our model, in which we develop the following experiment comparisons: 1) RGB and SRM: Xception-Base with single-modal input; 2) Two-stream: two-stream model with the direct summation of two modalities; 3) Two-att-stream: two-stream model with the attention-weighted module~\cite{Local-realtion}; 4) The proposed CMCE, LFGA, and MPFF modules of our model.

\par The experimental results are demonstrated in Tab \ref{tab6}. All models are trained on the FF++ dataset and evaluated on CD2 and DFD datasets. We make the following key observations: 1) The CMCE module outperforms the base two-stream model and the attention-based one. 2)The performance of our model has been incrementally reinforced by incorporating LFGA and MPFF modules. Overall, compared to the base two-stream model, our final model achieves an AUC improvement of $2.0\%$ and $0.9\%$, as well as an ACC improvement of $3.56\%$ and $1.66\%$ on the CD2 and DFD datasets, respectively.

\par We further develop the experiments regarding different facial regions chosen in the SSPSL strategy. The experimental results are demonstrated in Tab \ref{tab7}, in which the inner face is assigned as a rectangle formed by the boundaries of facial features, where nearly all pixels are manipulated in the FF++ dataset. We find that choosing the inner face as the reference region yields superior results, but it may contain more mistake real pixels when dealing with unknown datasets. Therefore, we ultimately choose the nose region to represent the manipulated face region’s distribution.

\section{Conclusion}
\par In the paper, we propose an innovative two-stream network, with remarkable generalization on unseen forgeries, that effectively considers the potential forged regions from which the model extracts adequate forgery evidence. We develop three novel modules CME, LFGA and MPFF to achieve our goals. For datasets without forgery annotations, we also propose a Semi-supervised Patch Similarity Learning strategy to adapt our model. Numerous experiments demonstrate that our method outperforms the best competitions on commonly used deepfake datasets, which indicates that our method can be a dependable solution in real-world scenarios to cope with the potential damages of deepfake.  

\bibliographystyle{ACM-Reference-Format}
\balance
\bibliography{refer}

\end{document}